\documentclass[sigconf]{acmart}

\usepackage{flushend}

\usepackage{booktabs} 

\setcitestyle{square}

\usepackage[ruled]{algorithm2e} 

\SetAlFnt{\small}
\SetAlCapFnt{\small}
\SetAlCapNameFnt{\small}
\SetAlCapHSkip{0pt}
\IncMargin{-\parindent}

\acmJournal{TOG}
\acmVolume{0}
\acmNumber{0}
\acmArticle{0}
\acmYear{2018}
\acmMonth{3}

\setcopyright{usgovmixed}

\acmDOI{}

\begin{document}
\title{Efficient Pose Tracking from Natural Features in Standard Web Browsers}
\author{Fabian G{\"o}ttl}
\affiliation{University of Passau}
\author{Philipp Gagel}
\affiliation{Coburg University of Applied Sciences and Arts}
\author{Jens Grubert}
\affiliation{Coburg University of Applied Sciences and Arts}
\email{jens.grubert@hs-coburg.de}



\begin{abstract}
Computer Vision-based natural feature tracking is at the core of modern Augmented Reality applications. Still, Web-based Augmented Reality typically relies on location-based sensing (using GPS and orientation sensors) or marker-based approaches to solve the pose estimation problem.

We present an implementation and evaluation of an efficient natural feature tracking pipeline for standard Web browsers using HTML5 and WebAssembly. Our system can track image targets at real-time frame rates tablet PCs (up to 60 Hz) and smartphones (up to 25 Hz). 

\end{abstract}

%
%


%
%


\maketitle

\section{Introduction}

We witness a proliferation of Augmented Reality (AR) applications such as games \cite{thomas2012survey} or utilities \cite{hartl2013mobile} across device classes (e.g., tablets, smartphones, smartglasses) and operating systems. Software Development Kits (SDKs) such as Vuforia, Apple ARKit, Google ARCore or the Windows Mixed Reality Toolit allow for efficient creation of AR applications for individual platforms through spatial tracking components with 6 degrees of freedom (DoF) pose estimation. However, deployment of AR solutions across platforms is often hindered by those platform-specific SDKs. 

Instead, existing Web-based AR solutions typically rely on location-based sensing (e.g., GPS plus orientation sensors).

In particular computer vision algorithms needed for realizing markerless Augmented Reality (AR) applications have been deemed too computational intensive for implementing them directly in the Web technology stack. While  marker-based AR systems (often derivates of ARToolkit) have been demonstrated to work in Web browsers \cite{ARjs}, natural feature tracking algorithms are not in widespread use on standard Web browsers. 

Recently, visual-inertial odometry approaches as provided by Apple ARKit or Google ARCore have been combined with custom Webviews \cite{arcoreweb, arkitweb} to allow experimentation with Web-based Cross Reality APIs like WebXR \cite{webxr}. Still, they are not available in standard Web browsers.

\begin{figure}[t!]
	\centering
	\includegraphics[width=\columnwidth]{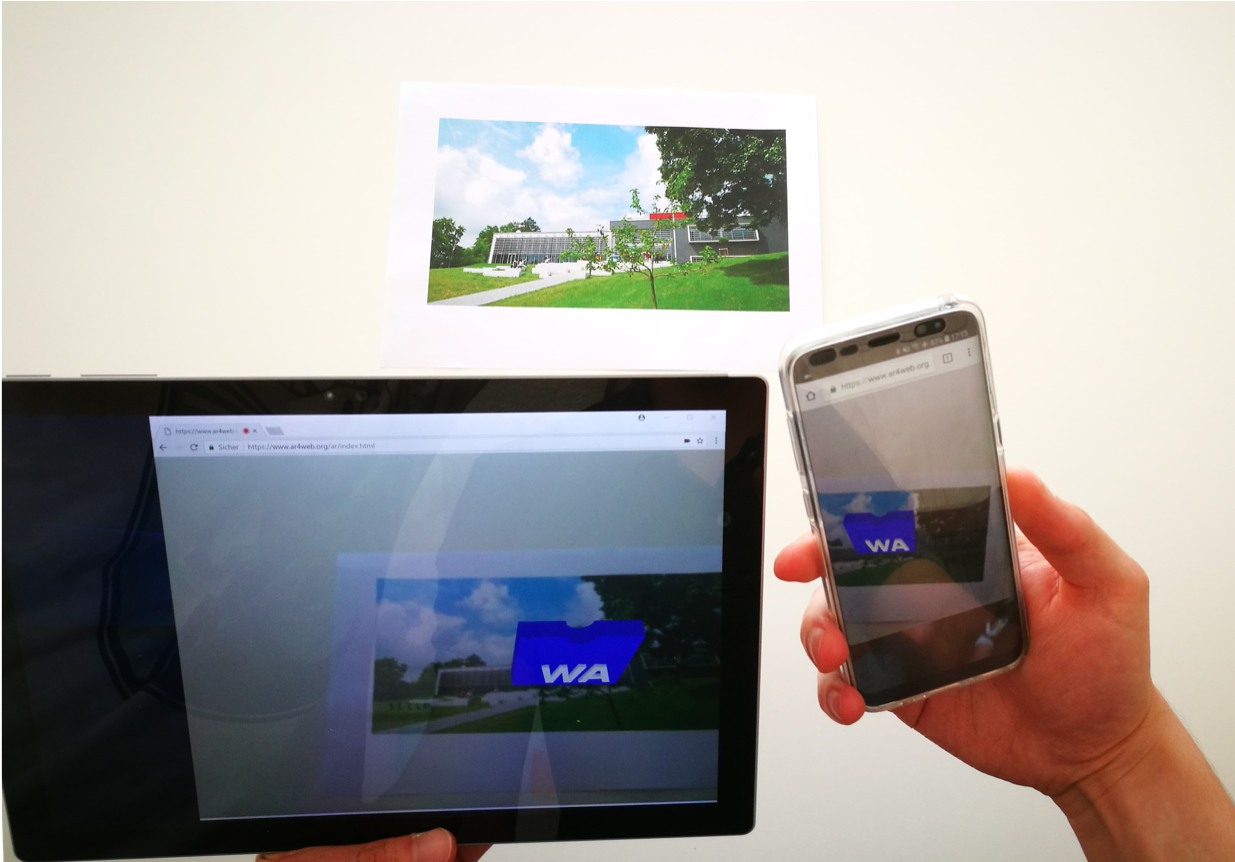}
	\caption{WebAssembly-based natural feature tracking pipeline running on Google Chrome on a Microsoft Surface Pro tablet and a Samsung Galaxy S8 smartphone.}
	\label{fig:s8example}
\end{figure}

Providing efficient 6 DoF pose estimation using natural features in standard Web browsers could help to overcome challenges of multiple creation of platform-specific code. To this end, we present the implementation and evaluation of an efficient computer vision-based pose estimation pipeline from natural features in standard Web browsers even on mobile devices, see Figure \ref{fig:s8example}. The pipeline can be tested under current versions of Google Chrome or Mozilla Firefox at www.ar4web.org.

\begin{figure*}[h!]
	\centering
	\includegraphics[width=2\columnwidth]{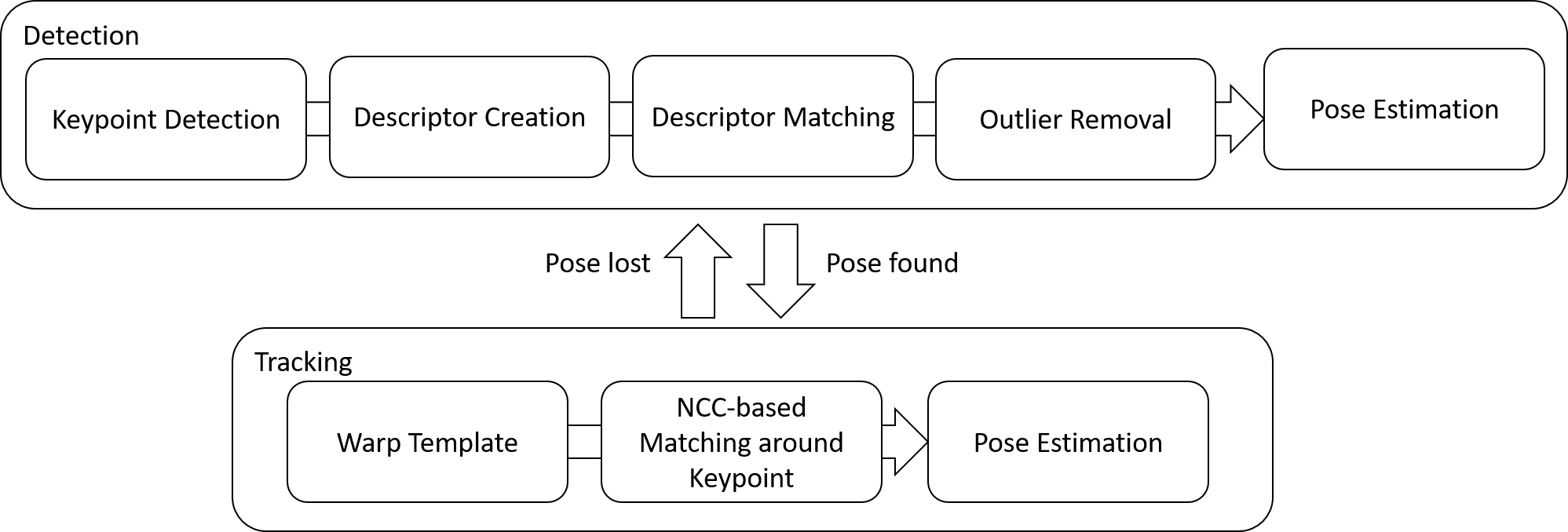}
	\caption{Overview of detection and tracking pipeline.}
	\label{fig:pipeline}
\end{figure*}

\section{Related Work}

There has been an ongoing interest in supporting the creation of Augmented Reality applications using Web technologies \cite{kooper2003browsing, macintyre2011argon, speiginer2015evolution}.

Several architectures have been proposed that aim at separating content, registration and presentation modules for Web-based Augmented Reality solutions (e.g., \cite{ahn2013webizing, ahn2014webizing, macintyre2011argon, leppanen2014augmented}) but they typically rely on sensor-based registration and tracking. Further, efforts have been made to standardize how to make available content into Web-based AR applications through XML-based formats (like ARML \cite{lechner2010arml, lechner2013arml}, KARML \cite{hill2010kharma} or the TOI format \cite{nixon2012smartreality}), which again focus on location-based spatial registration and often target mobile Augmented Reality browser applications \cite{sambinelli2015augmented, langlotz2013augmented, langlotz2014next}.

While recent experimental browsers have been released with visual-inertial tracking \cite{arcoreweb, arkitweb}, on standard Web browsers computer vision-based tracking is mainly limited to fidual-based pose estimation \cite{ARjs}.

Only few works have investigated to make computer-vision based pose tracking from natural features directly available using standard web technologies. In 2011, Oberhofer et al. presented an efficient natural feature tracking pipeline with a dedicated detection and tracking phase \cite{oberhofer2012natural}. While their solution was able to run in Web browsers supporting video capture through HTML5, the implementation was purely written in JavaScript. Specifically, it was not able to achieve real-time frame rates on mobile devices. They also reported that the slowdown compared to native implementations (in Google NativeClient) was fourfold.

Recently, a commercial solution for natural feature tracking was announced \cite{awe}. For this approach, no performance metrics were made available but only videos demonstrating interactive framerates on an unknown platform. 

In contrast to previous approaches, we present an efficient implementation of a pose estimation pipeline from natural features that runs at real-time framerates on standard web-browsers on mobile devices with processing time for the tracking component as low as 15 ms on a tablet and 50 ms on a smartphone.


\section{Natural Feature Tracking Pipeline}

Our pose estimation pipeline combines a dedicated detection step with an efficient tracking phase as proposed for mobile platforms \cite{wagner2008pose, wagner2010real}. Our overall pipeline is shown in Figure \ref{fig:pipeline}

If no pose has been found, the initial detection first extracts ORB features in the current camera frame \cite{rublee2011orb} and matches them with features from template images using fast approximate nearest neighbor search \cite{muja2009fast}. This is followed by outlier removal with a threshold of three times the minimum feature distance. Based on all remaining features, we first estimate a homography based on a RANSAC scheme, transform 4 corner points of the template image using that homography and employ an iterative perspective-n-point (PnP) algorithm for the final pose estimation.

As soon as the pose is found, we can switch from the expensive detection phase into a leightweight tracking phase. This phase consists of tracking keypoints that were detected previously in the detection phase. First, we take the homography H that was determined in the previous frame. Based on this homography, we create a warped representation of the marker. To speed up computation, the warped image is downsampled by factor of 2. For each keypoint that is visible in the current frame we cut a 5x5px image patch out of the warped image. The image patch should look similar to the patch in the current camera image. In order to save computation time, we keep track of a maximum of 25 patches or keypoints. With normalized cross correlation (NCC) we match the image patch to the current frame. Here, we use a search window of 16px length. These optimized points are stored with object keypoints together in tuples, which enables to compute an updated homography H. Similarly to the detection step, we compute the pose R and t out of H with iterative PnP. 

If the distance between the camera pose of the recent and previous frame is within a given threshold (e.g., the translation threshold for targets in DIN A4 was empirically determined as 5~cm), we handle the pose as valid.  At the next frame we start over again in front of this phase. If the pose is invalid, the detection phase is started again.

\subsection{Implementation}



Our pipeline utilizes WebAssembly and OpenCV for efficient processing of tracking data. We use the framework Emscripten, to compile C++ code with OpenCV embeddings to WebAssembly. Alternatively, Emscripten is able to output code in a subset of Javascript called asm.js. The camera is accessed through getUserMedia and writes the image into a HTML5 canvas element. The data is passed to the C++ level through heap memory and processed through the above mentioned pipeline. Finally, the computed pose is passed back and utilized in rendering a 3D scene with WebGL.

\begin{figure}[h!]
	\centering
	\includegraphics[width=\columnwidth]{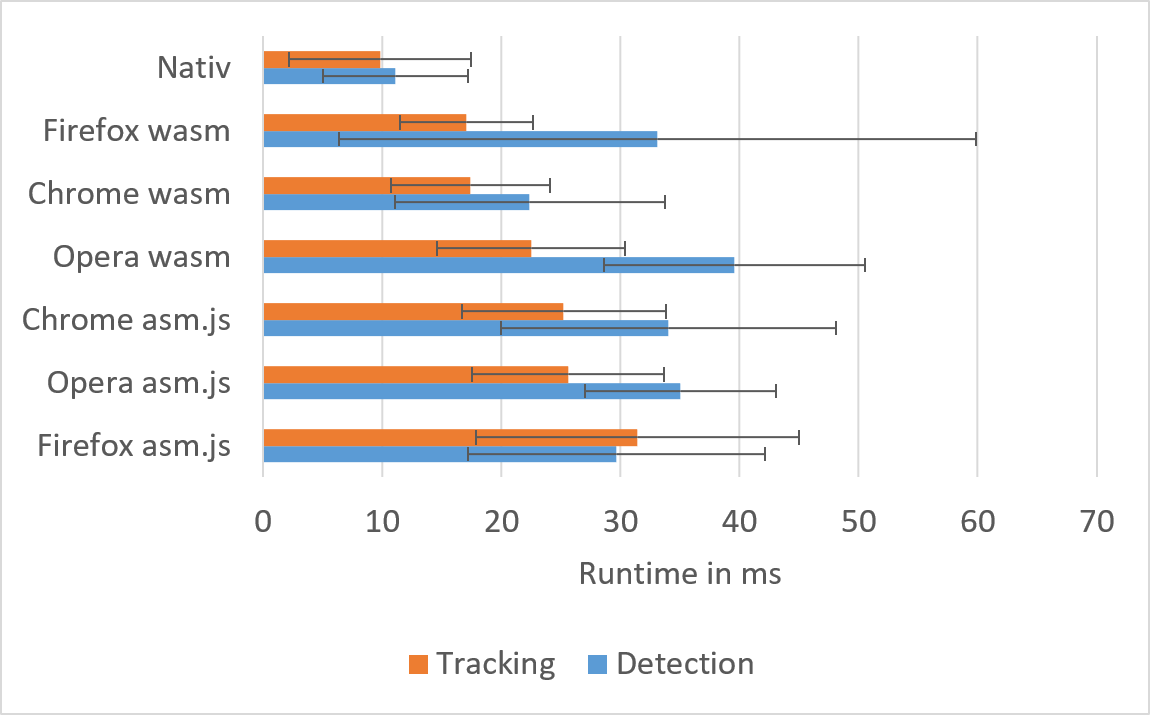}
	\caption{Runtime performance of the tracking and detection steps on a Microsoft Surface Pro.
	\\wasm: WebAssembly, asm.js: A low-level Javascript subset}
	\label{fig:rtsurface}
\end{figure}

\begin{figure}[h!]
	\centering
	\includegraphics[width=\columnwidth]{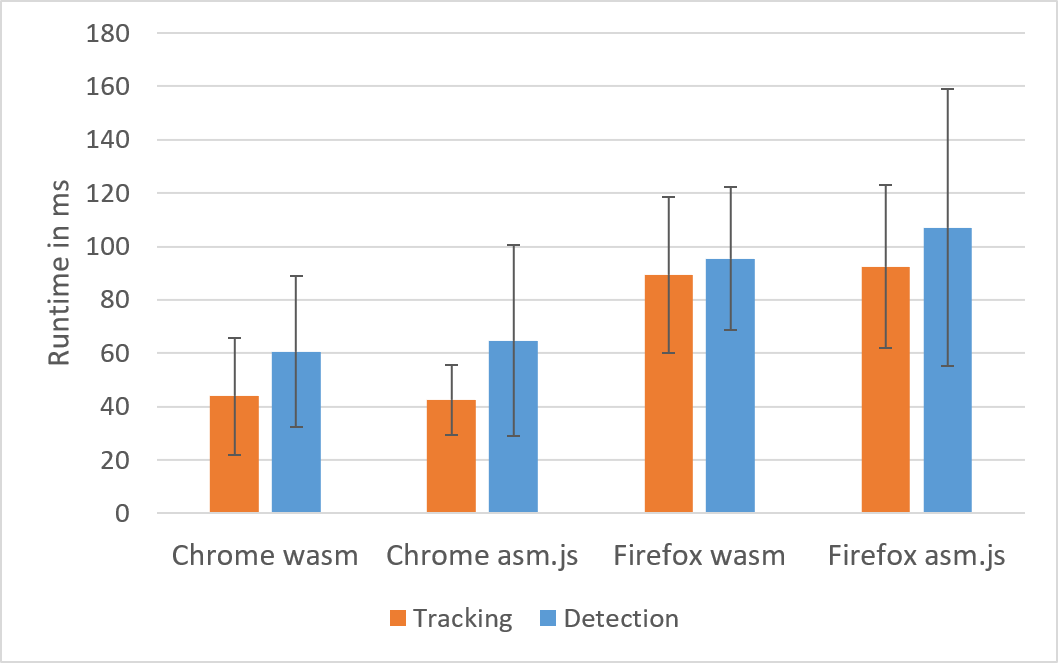}
	\caption{Runtime performance of the tracking and detection steps on a Samsung Galaxy S8.
	\\wasm: WebAssembly, asm.js: A low-level Javascript subset}
	\label{fig:rts8}
\end{figure}

\section{Evaluation}

We conducted a performance evaluation of the pipeline on a Tablet PC and a smartphone with two Web browsers on each platform (Mozilla Firefox version 59 and Google Chrome version 64). Additionally, we evaluated Opera 52 on the Tablet PC.

The Tablet PC was a Microsoft Surface Pro with an Intel Core i5-6300U (Dual Core with 2,40 Ghz) processor and 8 GB RAM running Windows 10. The smartphone was a Samsung Galaxy S8 with an Octa-core (4x2.3 GHz Mongoose M2 and 4x1.7 GHz Cortex-A53) processor and 4 GB RAM running Android 7.0 (Nougat). The camera resolution was set to 320x240 px. 

Figure \ref{fig:rtsurface} shows the runtime on the Tablet PC and Figure \ref{fig:rts8} for the smartphone. Please note, that in Figure \ref{fig:rtsurface}  there are additional bars indicating the performance of the native code. 


Compared to the native C++ implementation on a Microsoft Surface Pro the average runtime for detection increases by 100\% on Chrome to 200\% for Firefox. In contrast, for tracking the average runtime increases by 70\% for Firefox and by 75\% for Chrome.

\begin{figure}[h!]
	\centering
	\includegraphics[width=\columnwidth]{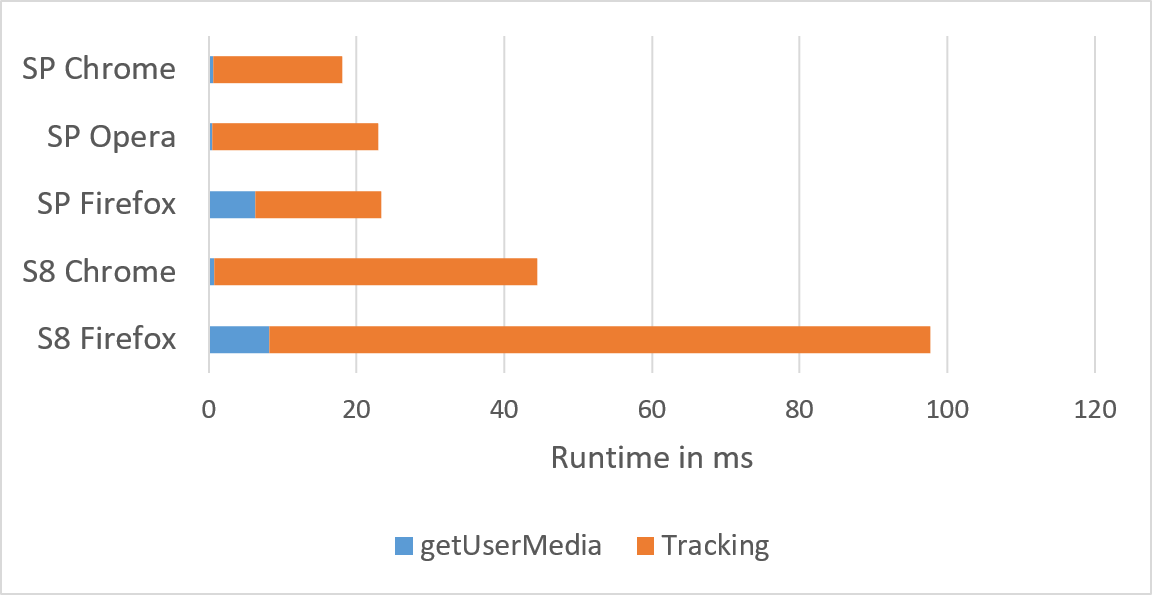}
	\caption{Runtime performance of the tracking pipeline including access to the camera image through getUserMedia. SP: Microsoft Surface Pro, S8: Samsung Galaxy S8.}
	\label{fig:rtcompletetracking}
\end{figure}

Figure \ref{fig:rtcompletetracking} indicates the average performance of the complete tracking pipeline (after an initial detection step). The figure indicates that accessing the camera through getUserMedia is substantially slower on Firefox compared to Google Chrome. Both on the Surface Pro and on the Galaxy S8 Google chrome needs on average 1~ms to deliver the camera image, whereas on Mozilla Firefox it is 6~ms on the Surface Pro and 8~ms on the Samsung Galaxy S8.

This indicates, that if the initial detection phase is completed, our pipeline can run up to 70 Hz under Chrome on a Surface Pro tablet and 20 Hz on a Galaxy S8. However, under Firefox it only runs at 30 Hz on a Surface Pro and 7 Hz on a Galaxy S8.

For robustness, we measured degrees until tracking failed. On multiple targets the minimum angle (starting from the horizontal plane) required for tracking was on average 17\textdegree (sd=3) both on the Tablet PC and the smartphone.


\section{Discussion}

The results indicate that under optimal conditions, the proposed pipeline can run efficiently on standard Web browsers, both on Tablet PCs as well as on recent smartphones. The initial detection step is rather slow (between 3 Hz on a smartphone and 12 Hz on a tablet PC) with a noticable speedup as soon as the tracking phase enters after the target was found initially. The real life runtime performance of the pipeline depends on how often a switch between both phases is necessary. We found empirically that even if the tracking phase fails the detection phase quickly re-initializes the pose and, hence, is only active for 1-2 frames.

We investigated the pipeline performance under WebAssembly and the JavaScript subset asm.js. WebAssembly builds are faster than asm.js. In contrast, WebAssembly builds are slower for the detection step in Firefox (by 9\%) and Opera (by 12\%). We assume that some optimizations for WebAssembly are not applicable in the detection step.

One issue we noticed during our evaluation, is the strong dependency of the pipeline runtime overall performance on the employed browser. Specifically, while Google Chrome can provide fast access to camera images in approximately 1~ms, Firefox can take up to 26~ms on a smartphone to access the image data.

To be fair, this issue is not specific to our implementation, but applies to other vision-based pipelines that require live camera access, as well.

\section{Conclusions and Future Work}

In this paper, we presented an implementation and evaluation of an efficient natural feature tracking pipeline for standard Web browsers using HTML5 and WebAssembly. Our system can track image targets at real-time frame rates tablet PCs (up to 65 Hz) and smartphones (up to 25 Hz). 

In future work, we want to combine our pipeline with efficient large scale image search and further optimize the tracking parameters (e.g., number of keypoints, search window size) on a per target basis. We also see potential for integrating it with Semantic Web-based Augmented Reality Systems \cite{nixon2012smartreality} or to utilize WebAssembly to enable new Web-based AR user experiences, e.g., through around-device interaction \cite{grubert2016glasshands}.

\bibliographystyle{ACM-Reference-Format}
\bibliography{web3d}

\end{document}